\documentclass[11pt,a4paper]{article}
\usepackage[hyperref]{acl2019}
\usepackage{times}
\usepackage{latexsym}

\usepackage{url}

\usepackage{graphicx}
\usepackage{amsfonts}
\usepackage{amsmath}
\usepackage{amssymb}
\usepackage{pifont}
\usepackage{nicefrac}
\usepackage{multirow}
\usepackage{subfigure}
\usepackage{caption}
\usepackage{todonotes}

\aclfinalcopy 

\title{A Lightweight Recurrent Network for Sequence Modeling}

\author{Biao Zhang$^1$ \quad Rico Sennrich$^{1,2}$  \bigskip\\
  $^1$School of Informatics, University of Edinburgh \\
  \texttt{B.Zhang@ed.ac.uk, rico.sennrich@ed.ac.uk} \medskip\\
    $^2$Institute of Computational Linguistics, University of Zurich}

\date{}

\begin{document}
\maketitle
\begin{abstract}
Recurrent networks have achieved great success on various sequential tasks with the assistance of complex recurrent units, but suffer from severe computational inefficiency due to weak parallelization. One direction to alleviate this issue is to shift heavy computations outside the recurrence. 
In this paper, we propose a lightweight recurrent network, or \textit{LRN}. 
LRN uses input and forget gates to handle long-range dependencies as well as gradient vanishing and explosion, with all parameter-related calculations factored outside the recurrence. The recurrence in LRN only manipulates the weight assigned to each token, tightly connecting LRN with self-attention networks. 
We apply LRN as a drop-in replacement of existing recurrent units in several neural sequential models. Extensive experiments on six NLP tasks show that LRN yields the best running efficiency with little or no loss in model performance.\footnote{Source code is available at \url{https://github.com/bzhangGo/lrn}.}
\end{abstract}

\section{Introduction}

Various natural language processing (NLP) tasks can be categorized as sequence modeling tasks, where recurrent networks (RNNs) are widely applied and contribute greatly to state-of-the-art neural systems~\cite{yang2017breaking,Peters:2018,8493282,P18-1008,kim2018semantic}. To avoid the optimization bottleneck caused by gradient vanishing and/or explosion~\cite{279181}, \citet{Hochreiter:1997:LSM:1246443.1246450} and \citet{D14-1179} develop gate structures to ease information propagation from distant words to the current position. Nevertheless, integrating these traditional gates inevitably increases computational overhead which is accumulated along token positions due to the sequential nature of RNNs. As a result, the weak parallelization of RNNs makes the benefits from improved model capacity expensive in terms of computational efficiency.

Recent studies introduce different solutions to this issue. \citet{D18-1459} introduce the addition-subtraction twin-gated recurrent unit (ATR), which reduces the amount of matrix operations by developing parameter-shared twin-gate mechanism. \citet{D18-1477} introduce the simple recurrent unit (SRU), which improves model parallelization by moving matrix computations outside the recurrence. Nevertheless, both ATR and SRU perform affine transformations of the previous hidden state for gates, though SRU employs a vector parameter rather than a matrix parameter. In addition, SRU heavily relies on its highway component, without which the recurrent component itself suffers from weak capacity and generalization~\cite{D18-1477}. 

In this paper, we propose a lightweight recurrent network (LRN), which combines the strengths of ATR and SRU. The structure of LRN is simple: an input gate and a forget gate are applied to weight the current input and previous hidden state, respectively. LRN has fewer parameters than SRU, and compared to ATR, removes heavy calculations outside the recurrence, generating gates based on the previous hidden state without any affine transformation. In this way, computation inside each recurrent step is highly minimized, allowing better parallelization and higher speed.

The gate structure endows LRN with the capability of memorizing distant tokens as well as handling the gradient vanishing and explosion issue. This ensures LRN's expressiveness and performance on downstream tasks. In addition, decomposing its recurrent structure discovers the correlation of input/forget gate with key/query in self-attention networks~\cite{NIPS2017_7181}, where these two gates together manipulate the weight assigned to each token. We also reveal how LRN manages long-term and short-term memories with the decomposition.

We carry out extensive experiments on six NLP tasks, ranging from natural language inference, document classification, machine translation, question answering and part-of-speech tagging to language modeling. We use LRN as a drop-in replacement of existing recurrent units in different neural models without any other modification of model structure. Experimental results show that LRN outperforms SRU by 10\%$\sim$20\% in terms of running speed, and is competitive with respect to performance and generalization compared against all existing recurrent units.

\section{Related Work}

Past decades have witnessed the rapid development of RNNs since the Elman structure was proposed~\cite{elman1990finding}. \citet{279181} point out that the gradient vanishing and explosion issue impedes the optimization and performance of RNNs. 
To handle this problem, \citet{Hochreiter:1997:LSM:1246443.1246450} develop LSTM where information and gradient from distant tokens can successfully pass through the current token via a gate structure and a memory cell. Unfortunately, the enhanced expressivity via complex gates comes at the cost of sacrificing computational efficiency, which becomes more severe when datasets are scaled up. Simplifying computation but keeping model capacity in RNNs raises a new challenge.

One direction is to remove redundant structures in LSTM. \citet{D14-1179} remove the memory cell and introduce the gated recurrent unit (GRU) with only two gates. \citet{lee2017recurrent} introduce an additive structure to generate hidden representations with linear transformed inputs directly, though we empirically observe that non-linear activations can stabilize model training. \citet{D18-1459} propose a twin-gate mechanism where input and forget gate are simultaneously produced from the same variables. We extend this mechanism by removing the affine transformation of previous hidden states.

Another direction is to shift recurrent matrix multiplications outside the recurrence so as to improve the parallelization of RNNs. \citet{bradbury2016quasi} propose the quasi-recurrent network (QRNN). QRNN factors all matrix multiplications out of the recurrence and employs a convolutional network to capture local input patterns. A minimal recurrent pooling function is used in parallel across different channels to handle global input patterns. \citet{Lei_etal_ICML2017} apply the kernel method to simplify recurrence and show improved model capacity with deep stacked RNNs. This idea is extended to SRU~\cite{D18-1477} where a minimal recurrent component is strengthened via an external highway layer. The proposed LRN falls into this category with the advantage over SRU of the non-dependence on the highway component.

Orthogonal to the above work, recent studies also show the potential of accelerating matrix computation with low-level optimization. \citet{diamos2016persistent} emphasize persistent computational kernels to exploit GPU's inverted memory hierarchy for reusing/caching purpose. \citet{appleyard2016optimizing} upgrade NIVIDIA's cuDNN implementation through exposing parallelism between operations within the recurrence. \citet{kuchaiev2017factorization} reduce the number of model parameters by factorizing or partitioning LSTM matrices. In general, all these techniques can be applied to any recurrent units to reduce computational overhead.

Our work is closely related with ATR and SRU. Although recent work shows that novel recurrent units derived from weighted finite state automata are effective without the hidden-to-hidden connection~\cite{Balduzzi:2016:SRN:3045390.3045527,D18-1152}, we empirically observe that including previous hidden states for gates is crucial for model capacity which also resonates with the evolution of SRU. Unlike ATR and SRU, however, we demonstrate that the affine transformation on the previous hidden state for gates is unnecessary. In addition, our model has a strong connection with self-attention networks.

\section{Lightweight Recurrent Network}

Given a sequence of input $\mathbf{X}$ = $[\mathbf{x}_1^\intercal; \mathbf{x}_2^\intercal; \ldots; \mathbf{x}_n^\intercal] \in \mathbb{R}^{n \times d}$ with length of $n$, LRN operates as follows\footnote{Bias terms are removed for clarity.}:
\begin{align}
    \mathbf{Q}, \mathbf{K}, \mathbf{V} = \mathbf{X}\mathbf{W}_q, \mathbf{X}\mathbf{W}_k, \mathbf{X}\mathbf{W}_v \label{eq_lrn_affine}\\
    \mathbf{i}_t = \sigma(\mathbf{k}_t + \mathbf{h}_{t-1}) \label{eq_lrn_igate}\\
    \mathbf{f}_t = \sigma(\mathbf{q}_t - \mathbf{h}_{t-1}) \label{eq_lrn_fgate}\\
    \mathbf{h}_t = g(\mathbf{i}_t \odot \mathbf{v}_t + \mathbf{f}_t \odot \mathbf{h}_{t-1}) \label{eq_lrn_out}
\end{align}
where $\mathbf{W}_q, \mathbf{W}_k, \mathbf{W}_v \in \mathbb{R}^{d \times d}$ are model parameters and $g(\cdot)$ is an activation function, such as \textit{identity} and \textit{tanh}. $\odot$ and $\sigma(\cdot)$ indicate the element-wise multiplication and sigmoid activation function, respectively. $\mathbf{q}_t, \mathbf{k}_t$ and $\mathbf{v}_t$ correspond to the $t$-th row of the projected sequence representation $\mathbf{Q}, \mathbf{K}, \mathbf{V}$. We use the term \textit{q}, \textit{k} and \textit{v} to denote the implicit correspondence to \textit{query}, \textit{key} and \textit{value} in self-attention networks which is elaborated in the next section.

As shown in Eq.~(\ref{eq_lrn_affine}), all matrix-related operations are shifted outside the recurrence and can be pre-calculated, thereby reducing the complexity of the recurrent computation from $\mathcal{O}(d^2)$ to $\mathcal{O}(d)$ and easing model parallelization. 
The design of the input gate $\mathbf{i}_t$ and forget gate $\mathbf{f}_t$ is inspired by the twin-gate mechanism in ATR~\cite{D18-1459}. Unlike ATR, however, we eschew the affine transformation on the previous hidden state. By doing so, the previous hidden state directly offers positive contribution to the input gate but negative to the forget gate, ensuring adverse correlation between these two gates.

The current hidden state $\mathbf{h}_t$ is a weighted average of the current input and the previous hidden state followed by an element-wise activation. When \textit{identity} function is employed, our model shows analogous properties to ATR. However, we empirically observe that this leads to gradually increased hidden representation values, resulting in optimization instability. Unlike SRU, which controls stability through a particular designed scaling term, we replace the \textit{identity} function with the \textit{tanh} function, which is simple but effective.

\section{Structure Decomposition}

In this section, we show an in-depth analysis of LRN by decomposing the recurrent structure. With an \textit{identity} activation, the $t$-th hidden state can be expanded as follows:
\begin{equation}\label{eq_str_dec}
    \begin{split}
        \mathbf{h}_t & = \sum_{k=1}^t \mathbf{i}_k \odot \left(\prod_{l=1}^{t-k}\mathbf{f}_{k+l}\right) \odot \mathbf{v}_k,
    \end{split}
\end{equation}
where the representation of the current token is composed of all previous tokens with their contribution distinguished by both input and forget gates. 

\textit{Relation with self-attention network.} 
After grouping these gates, we observe that:
\begin{equation}
        \mathbf{h}_t  = \sum_{k=1}^t \underbrace{\mathbf{i}_k}_{\text{key($\mathbf{K}$)}} \odot \underbrace{\mathbf{f}_{k+1} \odot \cdots \odot \mathbf{f}_t}_{\text{query($\mathbf{Q}$)}} \odot \underbrace{\mathbf{v}_k}_{\text{value($\mathbf{V}$)}}.
\end{equation}
Each weight can be regarded as a query from the current token $\mathbf{f}_t$ to the $k$-th input token $\mathbf{i}_k$. This query chain can be decomposed into two parts: a \textit{key} represented by $\mathbf{i}_k$ and a \textit{query} represented by $\prod_{l=1}^{t-k}\mathbf{f}_{k+l}$. 
The former is modulated through the weight matrix $\mathbf{W}_k$, and tightly associated with the corresponding input token. Information carried by the key remains intact during the evolution of time step $t$. In contrast, the latter, induced by the weight matrix $\mathbf{W}_q$, highly depends on the position and length of this chain, which dynamically changes between different token pairs.

The weights generated by keys and queries are assigned to \textit{value}s represented by $\mathbf{v}_k$ and manipulated by the weight matrix $\mathbf{W}_v$. Compared with self-attention networks, LRN shows analogous weight parameters and model structure. The difference is that weights in self-attention networks are normalized across all input tokens. Instead, weights in LRN are unidirectional, unnomalized and spanned over all channels.

\textit{Memory in LRN} 
Alternatively, we can view the gating mechanism in LRN as a memory that gradually forgets information.

Given the value representation at $k$-th time step $\mathbf{v}_k$, the information delivered to later time step $t$ ($k < t$) in LRN is as follows:
\begin{align}
    \underbrace{\mathbf{i}_k}_{\text{short term}} \odot \underbrace{\mathbf{f}_{k+1} \odot \cdots \odot \mathbf{f}_t}_{\text{forget chain (long term)}} \odot \mathbf{v}_k \label{eq_mem}.
\end{align}
The input gate $\mathbf{i}_k$ indicates the moment that LRN first accesses the input token $\mathbf{x}_k$, whose value reflects the amount of information or knowledge allowed from this token. A larger input gate corresponds to a stronger input signal, thereby a large change of activating short-term memory. This information is then delivered through a forget chain where memory is gradually decayed by a forget gate at each time step. 
The degree of memory decaying is dynamically controlled by the input sequence itself. When a new incoming token is more informative, the forget gate would increase so that previous knowledge is erased so as to make way for new knowledge in the memory. By contrast, meaningless tokens would be simply ignored.

\begin{table*}[t]
\begin{center}
{
\begin{tabular}{ll|c|c|c|c|c|c|c|c|c}
\multicolumn{2}{c|}{\multirow{2}{*}{Model}}  & \multirow{2}{*}{\#Params} & \multicolumn{2}{|c|}{Base} &  \multicolumn{2}{|c|}{+LN} &  \multicolumn{2}{|c|}{+BERT} &  \multicolumn{2}{|c}{+LN+BERT} \\
\cline{4-11}
& & & ACC & Time & ACC & Time & ACC & Time & ACC & Time \\
\hline
\multicolumn{2}{c|}{\citet{rocktaschel2016reasoning}} & 250K & 83.50 & - & - & - & - & - & - & -\\
\hline
\multicolumn{1}{c|}{\multirow{4}{*}{This}} 
& LSTM & 8.36M & 84.27 & 0.262 & 86.03 & 0.432 & 89.95 & 0.544 & \textbf{90.49} & 0.696 \\
\multicolumn{1}{l|}{}
& GRU & 6.41M & \textbf{85.71} & 0.245 & \textbf{86.05} & 0.419 & \textbf{90.29} & 0.529 & 90.10 & 0.695 \\
\multicolumn{1}{c|}{}
& ATR & 2.87M & 84.88 & 0.210 & 85.81 & 0.307 & 90.00 & 0.494 & 90.28 & 0.580 \\
\multicolumn{1}{c|}{Work}
& SRU & 5.48M & 84.28 & 0.258 & 85.32 & 0.283 & 89.98 & 0.543 & 90.09 & 0.555 \\
\cline{2-11}
\multicolumn{1}{l|}{}
& LRN & 4.25M & 84.88 & \textbf{0.209} & 85.06 & \textbf{0.223} & 89.98 & \textbf{0.488} & 89.93 & \textbf{0.506} \\ 
\end{tabular}
}
\end{center}
\caption{\label{tb_snli_result} Test accuracy (ACC) on SNLI task. ``\#Params'': the parameter number of Base. \textit{Base} and \textit{LN} denote the baseline model and layer normalization respectively. \textit{Time}: time in seconds per training batch measured from 1k training steps on GeForce GTX 1080 Ti. Best results are highlighted in bold.}
\end{table*}

\section{Gradient Analysis}

Gradient vanishing and explosion are the bottleneck that impedes training of vanilla RNNs~\cite{pascanu2013difficulty}. Consider a vanilla RNN formulated as follows:
\begin{equation}
    \mathbf{h}_t = g(\mathbf{W}\mathbf{x}_t + \mathbf{U}\mathbf{h}_{t-1}).
\end{equation}
The gradient back-propagated from the $t$-th step heavily depends on the following one-step derivation:
\begin{equation}\label{eq_der_rnn}
    \frac{\partial \mathbf{h}_t}{\partial \mathbf{h}_{t-1}} = \mathbf{U}^T g^{\prime}.
\end{equation}
Due to the \textit{chain rule}, the recurrent weight matrix $\mathbf{U}$ will be repeatedly multiplied along the sequence length. Gradient vanishing/explosion results from a weight matrix with small/large norm~\cite{pascanu2013difficulty}.

In LRN, however, the recurrent weight matrix is removed. The current hidden state is generated by directly weighting the current input and the previous hidden state. The one-step derivation of Eq.~(\ref{eq_lrn_igate}-\ref{eq_lrn_out}) is as follows:
\begin{equation}\label{eq_der_lrn}
    \frac{\partial \mathbf{h}_t}{\partial \mathbf{h}_{t-1}} = \underbrace{\left(\sigma_i^\prime \odot \mathbf{v}_t - \mathbf{h}_{t-1} \odot \sigma_f^\prime + \mathbf{f}_t \right)}_{\text{A}} \odot g^\prime
\end{equation}
where $\sigma_i^\prime$ and $\sigma_f^\prime$ denote the derivation of Eq.~(\ref{eq_lrn_igate}) and Eq.~(\ref{eq_lrn_fgate}), respectively. The difference between Eq.~(\ref{eq_der_rnn}) and Eq.~(\ref{eq_der_lrn}) is that the recurrent weight matrix is substituted by a more expressive component denoted as $\mathbf{A}$ in Eq.~(\ref{eq_der_lrn}). Unlike the weight matrix $\mathbf{U}$, the norm of $\mathbf{A}$ is input-dependent and varies dynamically along different positions. The dependence on inputs provides LRN with the capability of avoiding gradient vanishing/explosion.

\begin{table*}[t]
\begin{center}
{
\begin{tabular}{ll|c|c|c|c|c|c|c|c|c}
\multicolumn{2}{c|}{\multirow{2}{*}{Model}}  & \multirow{2}{*}{\#Params} & \multicolumn{2}{|c|}{AmaPolar} &  \multicolumn{2}{|c|}{Yahoo} &  \multicolumn{2}{|c|}{AmaFull} &  \multicolumn{2}{|c}{YelpPolar} \\
\cline{4-11}
& & & ERR & Time & ERR & Time & ERR & Time & ERR & Time \\
\hline
\multicolumn{2}{c|}{\citet{Zhang:2015:CCN:2969239.2969312}} & - & 6.10 & - & 29.16 & - & 40.57 & - & 5.26 & -\\
\hline
\multicolumn{1}{c|}{\multirow{4}{*}{This}} 
& LSTM & 227K & \textbf{4.37} & 0.947 & \textbf{24.62} & 1.332 & 37.22 & 1.003 & 3.58 & 1.362 \\
\multicolumn{1}{l|}{}
& GRU & 176K & 4.39 & 0.948 & 24.68 & 1.242 & \textbf{37.20} & 0.982  & \textbf{3.47} & 1.230 \\
\multicolumn{1}{c|}{}
& ATR & 74K  & 4.78 & 0.867 & 25.33 & 1.117 & 38.54 & 0.836 & 4.00 & 1.124 \\
\multicolumn{1}{c|}{Work}
& SRU & 194K & 4.95 & 0.919 & 24.78 & 1.394 & 38.23 & 0.907 & 3.99 & 1.310 \\
\cline{2-11}
\multicolumn{1}{l|}{}
& LRN & 151K  & 4.98 & \textbf{0.731} & 25.07 & \textbf{1.038} & 38.42 & \textbf{0.788} & 3.98 & \textbf{1.022} \\ 
\end{tabular}
}
\end{center}
\caption{\label{tb_doc_result} Test error (ERR) on document classification task. ``\#Params'': the parameter number in AmaPolar task. \textit{Time}: time in seconds per training batch measured from 1k training steps on GeForce GTX 1080 Ti.}
\end{table*}
\section{Experiments}

We verify the effectiveness of LRN on six diverse NLP tasks. For each task, we adopt (near) state-of-the-art neural models with RNNs handling sequence representation. We compare LRN with several cutting-edge recurrent units, including LSTM, GRU, ATR and SRU. For all comparisons, we keep the neural architecture intact and only alter the recurrent unit.\footnote{Due to possible dimension mismatch, we include an additional affine transformation on the input matrix for the highway component in SRU. In addition, we only report and compare speed statistics when all RNNs are optimally implemented where computations that can be done before the recurrence are moved outside.} All RNNs are implemented without specialized cuDNN kernels. Unless otherwise stated, different models on the same task share the same set of hyperparameters.

\subsection{Natural Language Inference}

\noindent \textbf{Settings} 
Natural language inference reasons about the entailment relationship between a premise sentence and a hypothesis sentence. We use the Stanford Natural Language Inference (SNLI) corpus~\cite{snli:emnlp2015} and treat the task as a three-way classification task. This dataset contains 549,367 premise-hypothesis pairs for training, 9,842 pairs for developing and 9,824 pairs for testing. We employ accuracy for evaluation.

We implement a variant of the word-by-word attention model~\cite{rocktaschel2016reasoning} using Tensorflow for this task, where we stack two additional bidirectional RNNs upon the final sequence representation and incorporate character embedding for word-level representation. The pretrained GloVe~\cite{pennington2014glove} word vectors are used to initialize word embedding. We also integrate the base BERT~\cite{devlin2018bert} to improve contextual modeling.

We set the character embedding size and the RNN hidden size to 64 and 300 respectively. Dropout is applied between consecutive layers with a rate of 0.3. We train models within 10 epochs using the Adam optimizer~\cite{kingma2014adam} with a batch size of 128 and gradient norm limit of 5.0. We set the learning rate to $1e^{-3}$, and apply an exponential moving average to all model parameters with a decay rate of 0.9999. These hyperparameters are tuned according to development performance.

\noindent \textbf{Results} 
Table \ref{tb_snli_result} shows the test accuracy and training time of different models. Our implementation outperforms the original model where \citet{rocktaschel2016reasoning} report an accuracy of 83.50. 
Overall results show that LRN achieves competitive performance but consumes the least training time. Although LSTM and GRU outperform LRN by 0.3$\sim$0.9 in terms of accuracy, these recurrent units sacrifice running efficiency (about 7\%$\sim$48\%) depending on whether LN and BERT are applied. No significant performance difference is observed between SRU and LRN, but LRN has fewer model parameters and shows a speedup over SRU of 8\%$\sim$21\%. 

Models with layer normalization (LN)~\cite{ba2016layer} tend to be more stable and effective. However, for LSTM, GRU and ATR, LN results in significant computational overhead (about 27\%$\sim$71\%). In contrast, quasi recurrent models like SRU and LRN only suffer a marginal speed decrease. This is reasonable because layer normalization is moved together with matrix multiplication out of the recurrence. 

Results with BERT show that contextual information is valuable for performance improvement. LRN obtains additional 4 percentage points gain with BERT and reaches an accuracy of around 89.9. This shows the compatibility of LRN with existing pretrained models. In addition, although the introduction of BERT brings in heavy matrix computation, the benefits from LRN do not disappear. LRN is still the fastest model, outperforming other recurrent units by 8\%$\sim$27\%.

\subsection{Document Classification}

\noindent \textbf{Settings} Document classification poses challenges in the form of long-range dependencies where information from distant tokens that contribute to the correct category should be captured. We use Amazon Review Polarity (AmaPolar, 2 labels, 3.6M/0.4M for training/testing), Amazon Review Full (AmaFull, 5 labels, 3M/0.65M for training/testing),  Yahoo! Answers (Yahoo, 10 labels, 1.4M/60K for training/testing) and Yelp Review Polarity (YelpPolar, 2 labels, 0.56M/38K for training/testing) from \citet{Zhang:2015:CCN:2969239.2969312} for experiments. We randomly select 10\% of training data for validation. Models are evaluated by test error.

We treat a document as a sequence of words. Our model is a bidirectional RNN followed by an attentive pooling layer. The word-level representation is composed of a pretrained GloVe word vector and a convolutional character vector. We use Tensorflow for implementation and do not use layer normalization. We set character embedding size to 32, RNN hidden size to 64 and dropout rate to 0.1. Model parameters are tuned by Adam optimizer with initial learning rate of $1e^{-3}$. Gradients are clipped when their norm exceeds 5. We limit the maximum document length to 400 and maximum training epochs to 6. Parameters are smoothed by an exponential moving average with a decay rate of 0.9999. These hyperparameters are tuned according to development performance.

\noindent \textbf{Results}
Table \ref{tb_doc_result} summarizes the classification results. LRN achieves comparable classification performance against ATR and SRU, but slightly underperforms LSTM and GRU (-0.45$\sim$-1.22). This indicates that LRN is capable of handling long-range dependencies though not as strong as complex recurrent units. Instead, the simplification endows LRN with less computational overhead than these units. Particularly, LRN accelerates the training over LSTM and SRU by about 20\%, or several days of training time on GeForce GTX 1080 Ti.\footnote{We notice that ATR operates faster than SRU. This is because though in theory SRU can be highly optimized for parallelization, computational framework like Tensorflow can not handle it automatically and the smaller amount of calculation in ATR has more advantage in practice.}

\begin{table}[t]
\centering
\begin{tabular}{l|c|c|c|c}
Model & \#Params & BLEU & Train & Decode \\
\hline
GNMT & - & 24.61 & - & - \\
\hline
GRU & 206M & 26.28 & 2.67 & 45.35 \\
ATR & 122M & 25.70 & 1.33 & \textbf{34.40} \\
SRU & 170M & 25.91 & 1.34 & 42.84 \\
\hline
LRN & 143M & 26.26 & \textbf{0.99} & 36.50 \\
oLRN & 164M & \textbf{26.73} & 1.15 & 40.19 
\end{tabular}
\caption{\label{tb_mt_result} Case-insensitive tokenized BLEU score on WMT14 English-German translation task. \textit{Train}: time in seconds per training batch measured from 0.2k training steps on Tesla P100. \textit{Decode}: time in milliseconds used to decode one sentence measured on newstest2014 dataset.} 
\end{table}
\subsection{Machine Translation}

\noindent \textbf{Settings} Machine translation is the task of transforming meaning from a source language to a target language. We experiment with the WMT14 English-German translation task~\cite{bojar-EtAl:2014:W14-33} which consists of 4.5M training sentence pairs.\footnote{Preprocessed data is available at~\cite{zhang-etal-2018-accelerating}: \url{https://drive.google.com/open?id=15WRLfle66CO1zIGKbyz0FsFmUcINyb4X}.} We use newstest2013 as our development set and newstest2014 as our test set. Case-sensitive tokenized BLEU score is used for evaluation.

We implement a variant of the GNMT system~\cite{45610} using Tensorflow, enhanced with residual connections, layer normalization, label smoothing, a context-aware component~\cite{Zhang:2017:CRE:3180104.3180106} and multi-head attention~\cite{NIPS2017_7181}. Byte-pair encoding~\cite{P16-1162} is used to reduce the vocabulary size to 32K. We set the hidden size and embedding size to 1024. Models are trained using Adam optimizer with adaptive learning rate schedule~\cite{P18-1008}. We cut gradient norm to 1.0 and set the token size to 32K. Label smoothing rate is set to 0.1.

\noindent \textbf{Model Variant}
Apart from LRN, we develop an improved variant for machine translation that includes an additional output gate. Formally, we change the Eq.~(\ref{eq_lrn_out}) to the following one:
\begin{align}
    \mathbf{c}_t = \mathbf{i}_t \odot \mathbf{v}_t + \mathbf{f}_t \odot \mathbf{h}_{t-1} \\
    \mathbf{o}_t = \sigma(\mathbf{W}_o \mathbf{x}_t - \mathbf{c}_t) \\
    \mathbf{h}_t = \mathbf{o}_t \odot \mathbf{c}_t
\end{align}
We denote this variant \textit{oLRN}. 
Like LRN, the added matrix transformation in oLRN can be shifted out of the recurrence, bringing in little computational overhead. The design of this output gate $\mathbf{o}_t$ is inspired by the LSTM structure, which acts as a controller to adjust information flow. In addition, this gate helps stabilize the hidden activation to avoid value explosion, and also improves model fitting capacity.  

\noindent \textbf{Results} 
The results in Table \ref{tb_mt_result} show that translation quality of LRN is slightly worse than that of GRU (-0.02 BLEU). After incorporating the output gate, however, oLRN yields the best BLEU score of 26.73, outperforming GRU (+0.45 BLEU). In addition, the training time results in Table \ref{tb_mt_result} confirm the computational advantage of LRN over all other recurrent units, where LRN speeds up over ATR and SRU by approximately 25\%. For decoding, nevertheless, the autoregressive schema of GNMT disables position-wise parallelization. In this case, the recurrent unit with the least computation operations, i.e. ATR, becomes the fastest. Still, both LRN and oLRN translate sentences faster than SRU (+15\%/+6\%).

\begin{table}[t]
\centering
\begin{tabular}{l|c|c|c}
{Model} & {\#Params} & {Base} & {+Elmo} \\
\hline
rnet* & - & 71.1/79.5 & -/- \\
\hline
LSTM & 2.67M & \textbf{70.46}/78.98 & 75.17/82.79 \\
GRU & 2.31M & 70.41/\textbf{79.15} & 75.81/83.12 \\
ATR & 1.59M & 69.73/78.70 & 75.06/82.76 \\
SRU & 2.44M & 69.27/78.41 & 74.56/82.50 \\
\hline
LRN & 2.14M & 70.11/78.83 & \textbf{76.14}/\textbf{83.83} \\ 
\end{tabular}
\caption{\label{tb_qa_result} Exact match/F1-score on SQuad dataset. ``\#Params'': the parameter number of Base. \textit{rnet*}: results published by~\citet{P17-1018}.} 
\end{table}

\begin{table*}[t]
\begin{center}
{
\begin{tabular}{ll|c|c|c|c||c|c|c}
\multicolumn{2}{c|}{\multirow{2}{*}{Model}}  & \multirow{2}{*}{\#Params} & \multicolumn{3}{|c||}{PTB} &  \multicolumn{3}{c}{WT2}  \\
\cline{4-9}
& & & Base & +Finetune & +Dynamic & Base & +Finetune & +Dynamic \\
\hline
\multicolumn{2}{c|}{\citet{yang2017breaking}} & 22M & 55.97 & 54.44 & 47.69 & 63.33 & 61.45 & 40.68 \\
\hline
\multicolumn{1}{c|}{\multirow{4}{*}{This}} 
& LSTM & 22M & 63.78 & 62.12 &  \textbf{53.11} & \textbf{69.78} & \textbf{68.68} & \textbf{44.60} \\
\multicolumn{1}{l|}{}
& GRU & 17M & 69.09 & 67.61 & 60.21 & 73.37 & 73.05 & 49.77 \\
\multicolumn{1}{c|}{}
& ATR & 9M & 66.24 & 65.86 & 58.29 & 75.36 & 73.35 & 48.65 \\
\multicolumn{1}{c|}{Work}
& SRU & 13M & 69.64 & 65.29 & 60.97 & 85.15 & 84.97 & 57.97 \\
\cline{2-9}
\multicolumn{1}{l|}{}
& LRN & 11M & \textbf{61.26} & \textbf{61.00} & 54.45 & 69.91 & 68.86 & 46.97 \\ 
\end{tabular}
}
\end{center}
\caption{\label{tb_lm_result} Test perplexity on PTB and WT2 language modeling task. ``\#Params'': the parameter number in PTB task. \textit{Finetune}: fintuning the model after convergence. \textit{Dynamic} dynamic evaluation. Lower perplexity indicates better performance.}
\end{table*}

\begin{table}[t]
\centering
\begin{tabular}{l|c|c}
{Model} & {\#Params} & {NER} \\
\hline
LSTM* & - & 90.94 \\
\hline
LSTM & 245K & \textbf{89.61} \\
GRU & 192K  & 89.35  \\
ATR & 87K  & 88.46 \\
SRU &  161K & 88.89 \\
\hline
LRN & 129K & 88.56\\ 
\end{tabular}
\caption{\label{tb_ner_result} F1 score on CoNLL-2003 English NER task.  ``\#Params'': the parameter number in NER task. \textit{LSTM*} denotes the reported result~\cite{N16-1030}.}
\end{table}
\subsection{Reading Comprehension}

\noindent \textbf{Settings} Reading comprehension aims at providing correct answers to a query based on a given document, which involves complex sentence matching, reasoning and knowledge association. We use the SQuAD corpus~\cite{D16-1264} for this task and adopt span-based extraction method. This corpus contains over 100K document-question-answer triples. We report exact match (EM) and F1-score (F1) on the development set for evaluation.

We employ the public available rnet model~\cite{P17-1018}\footnote{\url{https://github.com/HKUST-KnowComp/R-Net}} in Tensorflow. We use the default model settings: character embedding size 8, hidden size 75, batch size 64, and Adadelta optimizer~\cite{zeiler2012adadelta} with initial learning rate of 0.5. Gradient norm is cut to 5.0. We also experiment with Elmo~\cite{Peters:2018}, and feed the Elmo representation in before the encoding layer and after the matching layer with a dropout of 0.5.

\noindent \textbf{Results} Table \ref{tb_qa_result} lists the EM/F1 score of different models. In this task, LRN outperforms ATR and SRU in terms of both EM and F1 score. After integrating Elmo for contextual modeling, the performance of LRN reaches the best (76.14 EM and 83.83 F1), beating both GRU and LSTM (+0.33EM, +0.71F1). As recent studies show that cases in SQuAD are dominated by local pattern matching~\cite{D17-1215}, we argue that LRN is good at handling local dependencies.

\subsection{Named Entity Recognition}

\noindent \textbf{Settings} Named entity recognition (NER) classifies named entity mentions into predefined categories. We use the CoNLL-2003 English NER dataset~\cite{TjongKimSang:2003:ICS:1119176.1119195} and treat NER as a sequence labeling task. We use the standard train, dev and test split. F1 score is used for evaluation.

We adopt the bidirectional RNN with CRF inference architecture~\cite{N16-1030}. We implement different models based on the public codebase in Tensorflow.\footnote{\url{https://github.com/Hironsan/anago}} We use the default hyperparameter settings. Word embedding is initialized by GloVe vectors.

\begin{figure*}
\centering
\includegraphics[scale=0.32]{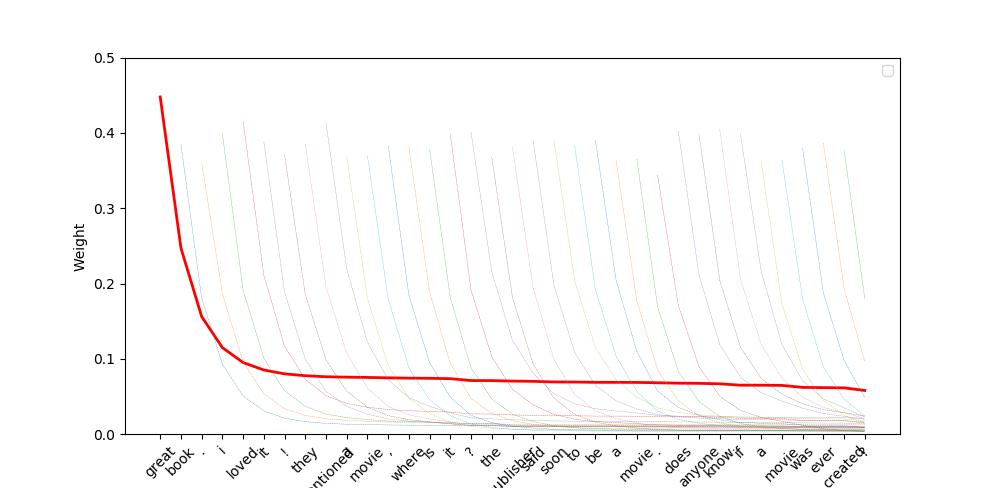}
\caption{\label{fig_mem} The decay curve of each token modulated by input and forget gates along the token position. Notice how the memory of term ``\textit{great}'' flows to the final state shown in red, and contributes to a \textit{Positive} decision. \textit{Weight} denotes the averaged activation of $\mathbf{i}_k\odot \left(\prod_{l=1}^{t-k}\mathbf{f}_{k+l}\right)$ as shown in Eq.~(\ref{eq_str_dec}).}
\end{figure*}

\noindent \textbf{Results} As shown in Table \ref{tb_ner_result}\footnote{Notice that our implementation falls behind the original model~\cite{N16-1030} because we do not use specifically trained word embedding.}, the performance of LRN matches that of ATR and SRU, though LSTM and GRU operate better (+1.05 and +0.79). As in the SQuAD task, the goal of NER is to detect local entity patterns and figure out the entity boundaries. However, the performance gap between LSTM/GRU and LRN in NER is significantly larger than that in SQuAD. We ascribe this to the weak model architecture and the small scale NER dataset where entity patterns are not fully captured by LRN.

\subsection{Language Modeling}

\noindent \textbf{Settings} Language modeling aims to estimate the probability of a given sequence, which requires models to memorize long-term structure of language. We use two widely used datasets, Penn Treebank (PTB)~\cite{mikolov2010recurrent} and WikiText-2 (WT2)~\cite{merity2016pointer} for this task. Models are evaluated by perplexity.

We modify the mixture of softmax model (MoS)~\cite{yang2017breaking}\footnote{\url{https://github.com/zihangdai/mos}} in PyTorch to include different recurrent units. We apply weight dropout to all recurrent-related parameters instead of only hidden-to-hidden connection. We follow the experimental settings of MoS, and manually tune the initial learning rate based on whether training diverges. 

\noindent \textbf{Results} Table \ref{tb_lm_result} shows the test perplexity of different models.\footnote{Our re-implementation of LSTM model is worse than the original model~\cite{yang2017breaking} because the system is sensitive to hyperparameters, and we apply weight dropout to all LSTM parameters which makes the original best choices not optimal.} In this task, LRN significantly outperforms GRU, ATR and SRU, and achieves near the same perplexity as LSTM. This shows that in spite of its simplicity, LRN can memorize long-term language structures and capture a certain degree of language variation. In summary, LRN generalizes well to different tasks and can be used as a drop-in replacement of existing recurrent units.

\subsection{Ablation Study}

\begin{table}[t]
\centering
\begin{tabular}{l|c|c}
{Model} & {SNLI} & {PTB} \\
\hline
\hline
LRN & \textbf{85.06} & \textbf{61.26} \\
gLRN & 84.72 &  92.49 \\
eLRN & 83.56 & 169.81 \\
\end{tabular}
\caption{\label{tb_abs_result} Test accuracy on SNLI task with \textit{Base+LN} setting and test perplexity on PTB task with \textit{Base} setting.}
\end{table}

Part of LRN can be replaced with some alternatives. In this section, we conduct ablation analysis to examine two possible designs:
\begin{description}
    \item[\normalfont \textit{gLRN}] The twin-style gates in Eq.~(\ref{eq_lrn_igate}-\ref{eq_lrn_fgate}) can be replaced with a general one: 
    \begin{equation}
        \mathbf{f}_t = \sigma(\mathbf{q}_t - \mathbf{h}_{t-1}), \mathbf{i}_t = 1 - \mathbf{f}_t.
    \end{equation}
    In this way, input and forget gate are inferable from each other with the key weight parameter removed.
    \item[\normalfont \textit{eLRN}] The above design can be further simplified into an extreme case where the forget gate is only generated from the previous hidden state without the query vector: 
    \begin{equation}
        \mathbf{f}_t = \sigma( - \mathbf{h}_{t-1}), \mathbf{i}_t = 1 - \mathbf{f}_t.
    \end{equation}
\end{description}
We experiment with SNLI and PTB tasks. Results in Table \ref{tb_abs_result} show that although the accuracy on SNLI is acceptable, gLRN and eLRN perform significantly worse on the PTB task. This suggests that these alternative structures suffer from weak generalization.

\subsection{Structure Analysis}

In this section, we provide a visualization to check how the gates work in LRN. 

We experiment with a unidirectional LRN on the AmaPolar dataset, where the last hidden state is used for document classification. Figure \ref{fig_mem} shows the decay curve of each token along the token position. The memory curve of each token decays over time. However, important clues that contribute significantly to the final decision, as the token ``\textit{great}'' does, decrease slowly, as shown by the red curve. Different tokens show different decay rate, suggesting that input and forget gate are capable of learning to propagate relevant signals. All these demonstrate the effectiveness of our LRN model.

\section{Conclusion and Future Work}

This paper presents LRN, a lightweight recurrent network that factors matrix operations outside the recurrence and enables higher parallelization. Theoretical and empirical analysis shows that the input and forget gate in LRN can learn long-range dependencies and avoid gradient vanishing and explosion.
LRN has a strong correlation with self-attention networks.
Experiments on six different NLP tasks show that LRN achieves competitive performance against existing recurrent units. It is simple, effective and reaches better trade-off among parameter number, running speed, model performance and generalization.

In the future, we are interested in testing low-level optimizations of LRN, which are orthogonal to this work, such as dedicated cuDNN kernels.

\section*{Acknowledgments}

We thank the reviewers for their insightful comments.
Biao Zhang acknowledges the support of the Baidu Scholarship. This work has been performed using resources provided by the Cambridge Tier-2 system operated by the University of Cambridge Research Computing Service (http://www.hpc.cam.ac.uk) funded by EPSRC Tier-2 capital grant EP/P020259/1.

\bibliography{acl2019}
\bibliographystyle{acl_natbib}

\end{document}